\documentclass[letterpaper, 10 pt, conference]{ieeeconf}  

\IEEEoverridecommandlockouts                              

\usepackage{color}
\usepackage{cite}
\usepackage{graphics} 
\usepackage{epsfig} 
\usepackage{mathptmx} 
\usepackage{times} 
\usepackage{amsmath} 
\usepackage{amssymb}  
\usepackage[caption=false,font=normalsize,labelfont=sf,textfont=sf]{subfig}
\usepackage{booktabs} 
\usepackage[colorlinks,linkcolor=red]{hyperref}

\title{\LARGE \bf
Tightly Joining Positioning and Control for Trustworthy Unmanned Aerial Vehicles Based on Factor Graph Optimization in Urban Transportation }

\author{Peiwen Yang and Weisong Wen
\thanks{\color{black}{This work was supported by funded by the Guangdong Basic and Applied Basic Research Foundation (2021A1515110771) and University Grants Committee of Hong Kong under the scheme Research Impact Fund (R5009-21). This research was also supported by the Faculty of Engineering, The Hong Kong Polytechnic University under the project ``Perception-based GNSS PPP-RTK/LVINS integrated navigation system for unmanned autonomous systems operating in urban canyons".}}
\thanks{Weisong, Wen is the corresponding author. The authors are with the Department of Aeronautical and Aviation Engineering,
        The Hong Kong Polytechnic University, HongKong, China (e-mail: welson.wen@polyu.edu.hk).}
}

\begin{document}

\maketitle
\thispagestyle{empty}
\pagestyle{empty}

\begin{abstract}
Unmanned aerial vehicles (UAV) showed great potential in improving the efficiency of parcel delivery applications in the coming smart cities era. Unfortunately, the trustworthy positioning and control algorithms of the UAV are significantly challenged in complex urban areas. For example, the ubiquitous global navigation satellite system (GNSS) positioning can be degraded by the signal reflections from surrounding high-rising buildings, leading to significantly increased positioning uncertainty. An additional challenge is introduced to the control algorithm due to the complex wind disturbances in urban canyons. Given the fact that the system positioning and control are highly correlated with each other, for example, the system dynamics of the control can largely help with the positioning, this paper proposed a joint positioning and control method (JPCM) based on factor graph optimization (FGO), which combines sensors' measurements and control intention. In particular, the positioning measurements are formulated as the factors in the factor graph model, such as the positioning from the GNSS. The model predictive control (MPC) is also formulated as the additional factors in the factor graph model. By solving the factor graph contributed by both the positioning factor and the MPC-based factors, the complementariness of positioning and control can be fully explored. To guarantee reliable system dynamic parameters, we validate the effectiveness of the proposed method using a simulated quadrotor system which showed significantly improved trajectory following performance. To benefit the research community, we open-source our code and make it available at \url{https://github.com/RoboticsPolyu/IPN\_MPC}. 
\end{abstract} 

\section{INTRODUCTION}
\begin{figure}[th]
      \centering
      \includegraphics[width=0.8\linewidth]{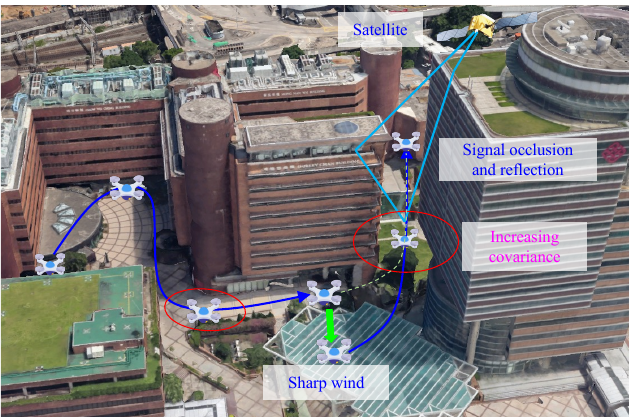}
      \caption{The safety challenges of intelligent transportation in smart cities (the background figure is from Google Earth \cite{Google}).}
      \label{img:urban_city}
 \end{figure}

UAV is important for parcel delivery but is challenging in urban canyons. As shown in Fig. \ref{img:urban_city}, occlusion and reflection of GNSS signals can increase the positioning uncertainty due to surrounding high-rising buildings \cite{HuangFeng,Outliers}, which poses risks to UAV's control. Additionally, unexpected movements caused by sharp wind or other reasons may result in unforeseeable planning actions of UAVs. {\color{black}{The biological} control system appears to incorporate both positioning and control mechanisms, thereby ensuring the safety of control operations even in the presence of high levels of positioning noise. }The FGO has been widely applied in the field of urban vehicles positioning \cite{RN341,RN378, WenLiDARAided} and optimization-based control problem \cite{Yang_2021, xie2020factorgraph, RN812, ChenCableRobot}. In addition, trajectory estimation and planning can be combined and showed some improvement compared with the conventional methods \cite{RN838,SCTEforCollision}. This is because the factor graphs solving optimization problems naturally allow unifying the positioning and control under the same mathematics representation. Therefore, we propose a JPCM that integrates measurement factors and control-related factors into a unified factor graph. 

FGO has been emerging in the field of planning and control. Equality constrained linear optimal control was handled by factor graphs \cite{Yang_2021}. Moreover, Xie \cite{xie2020factorgraph} introduced both equality and inequality constraints in factor graphs. Then, bazzana \cite{RN812} modelled optimal control for local planning with factor graphs. Chen \cite{ChenCableRobot} applied FGO to generate a nominal trajectory for cable-driven parallel robots, then linearized the graph and used variable elimination to compute the feedback gains. Besides, mukadam \cite{RN838} proposed the STEAP method which combined trajectory estimation and path planning with obstacle avoidance factors and goal factor. But, the method was un-adapt to trajectory tracking. And Matthew \cite{SCTEforCollision} extended STEAP to dynamic environments with moving obstacles.  Besides, FGO can be applied to solve kinematic and dynamics problems for multibody systems \cite{MultiBodyFGO}.  The above papers revealed the feasibility of applying factor graphs to trajectory planning or control. Notwithstanding, the real-time control method that integrates positioning and control, especially for high-speed and highly nonlinear UAV, remains a research hotspot.

UAVs are versatile since they allow vertical take-off and landing, great manoeuvrability, and hovering tasks \cite{RN771,SurveyMAV, EGO}. MPC \cite{SurveyOnPF} has been extensively studied in the field of intelligent vehicles \cite{ITSC_MPCV, ITSC_PC_MPC, ITSC_SC}. Moreover, many researchers have investigated the MPC-based UAV application in challenging control problems. To avoid over-parameterization or singularity, on-manifold representation MPC was proposed in \cite{RN441}. Romero \cite{RN87, Time-Optimal} solved the time optimal control problem via model predictive contouring control. In addition, a UAV's fault-tolerant controller was presented based on nonlinear model predictive control (NMPC) when the complete failure of a single rotor occurred \cite{RN89}. Lindqvist \cite{CAMPC} proposed a baseline controller for collision avoidance with dynamic obstacles. Wang \cite{EfficientMPC} contributed on how to solve the heavy computational complexity problem of NMPC. 
Moreover, some robust MPC methods that handled the dynamics system with unknown but bounded disturbance, such as Min-max MPC, tube MPC \cite{ITSC_TUBE}, etc \cite{RN871}. The algorithms mentioned above ignored the uncertainty of positioning in urban scenarios. On the contrary, joint optimization can directly derive control laws by introducing measurement constraints into control problems. 




\begin{figure}[ht]
      \centering
      \includegraphics[width=0.8\linewidth]{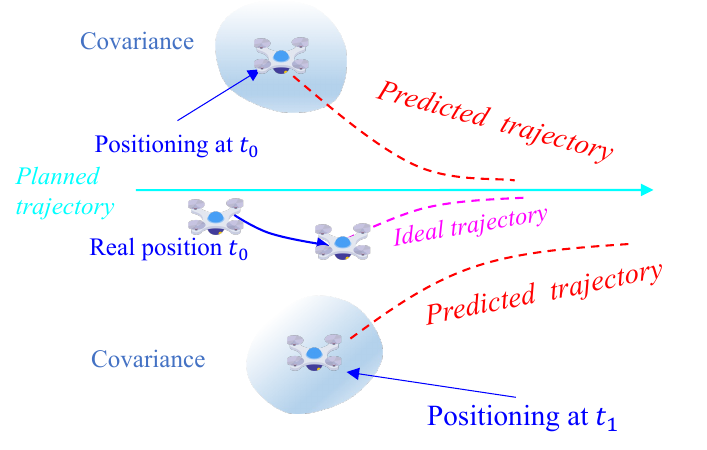}
      \caption{The control jitter due to UAV's positioning uncertainty.}
      \label{img:ProblemIntro}
\end{figure}

\begin{figure}[ht]
      \centering
      \includegraphics[width=0.9\linewidth]{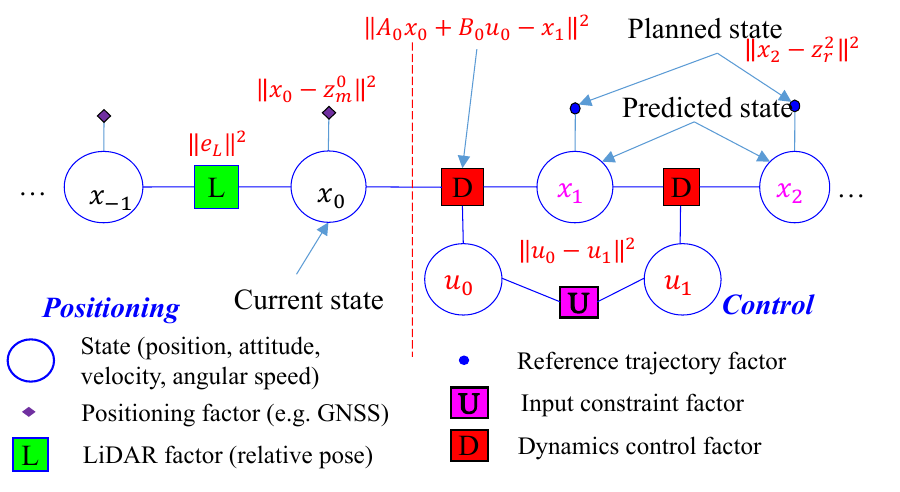}
      \caption{The factor graph overview of joint positioning and control method.}
      \label{img:FGIntroSmall}
\end{figure}

As depicted in Fig. \ref{img:ProblemIntro}, the nominal MPC attempts to compute a feasible solution that enables a predicted trajectory (red dashed line) to approach the planned trajectory as closely as possible. However, the predicted trajectory may fluctuate heavily from $t_0$ to $t_1$ due to positioning error, leading to the application of input commands that do not match the real UAV's state. Hence, the control stability under positioning errors is essential for safety.  In this work, as illustrated in Fig. \ref{img:FGIntroSmall}, the current state is connected with the predicted state by a dynamics control factor, which plays the role of a bridge between positioning and control.

This paper proposes a positioning and control joint optimization model by combining positioning and control into a unified factor graph. The problem is solved iteratively, which simultaneously integrates the constraints of sensors' measurements and planned trajectory. In addition, the measurement constraints include absolute positioning constraint (e.g. GNSS positioning factor) and relative pose constraint (e.g. LiDAR factor). Simulation results demonstrate that the proposed JPCM achieves more accuracy control compared with nominal MPC.   Additionally, it also effectively suppresses the attitude jitter caused by large positioning errors.

\section{PROBLE DESCRIPTION}

In this paper, the coordinate system consists of the world coordinate system $F_w$, the UAV body coordinate system $F_b$, and the LiDAR coordinate system ${F_L}$. The extrinsic parameter from the UAV body coordinate system to the LiDAR coordinate system is $T_b^L$.

\subsection{UAV Dynamics model}

The generalization dynamics model of a UAV is as follows:
\begin{subequations}
	\begin{align}
			\dot{x} &= f(x,\tau) \label{eq:Dyn} \\
			\tau &= g(u),  \label{eq:INPUT}
	\end{align}
\end{subequations}
where $x = [p^w_b, R^w_b, v^w_b, \omega_b]$ is the UAV's state that includes position $p^w_b$, rotation $R^w_b$, velocity $v^w_b$, and angular speed $\omega_b$, and $\tau = [T_b,  M_b]$ represents the resultant thrust $T_b \in R^3$ and resultant moments $M_b \in R^3$ generated by the UAV's rotors, and $u = [u_0, ... , u_{K-1}]^T, K > 0, K \in N$ is the control input vector. For example, $u_j$ represents the revolutions per minute (RPM) of the $j$th motor of a quadrotor. $f(*)$ and $g(*)$ represent the generalization dynamics model and control allocation model, respectively. 

The continuous-time linearized dynamics of (\ref{eq:Dyn}) is as follows:
\begin{subequations}
	\begin{align}
		\dot{p}^w_b &= v^w_b \\
		\dot{R}^w_b &= {\color{black}R^w_b}  \lfloor \omega_b \times \rfloor \\
		m_b \dot{v}^w_b &= - m_b e_3 g + R^w_bT_b \\
		I_b \dot{\omega}_b &=-\omega_b \times I_b \omega_b + M_b,
	\end{align}\label{eq:CLD_Dyn}
\end{subequations}
where $m_b$ is UAV's mass, $e_3 = [0,0,1]^T$, $g$ is gravity constant, and $\lfloor * \times \rfloor$ represents skew-symmetric matrix operator. And $I_b$ is the moment of inertia, which is usually regarded as a diagonal matrix:
\begin{equation}
	I_b = diag([I^1_b, I^2_b, I^3_b]),
\end{equation} 
where $diag$ is a is an operation that converts a vector into a diagonal matrix.

Moreover, the symbol $Q_k, Q_N, R_t$ should also be further explained, especially its relationship to $W_k, Y_k$ in (7).


The multi-rotor UAV is driven by $K$ propellers that rotate around the motor's shaft. Every rotor can generate corresponding thrust and moments. The model of thrust and moments is as follows: 
\begin{subequations}
	\begin{align}
		T_b &=\sum^{K-1}_{j=0} axis^b_j \cdot F^j_b = \sum^{K-1}_{j=0} {\color{blue}  axis^b_j \cdot c_t}\cdot A_j  = \sum^{K-1}_{j=0} f_j A_j \label{eq:TRUST} \\
		\begin{split}
					M_b &= \sum^{K-1}_{j=0} {M^j_b} = \sum^{K-1}_{j=0} {\color{blue} (P^b_{r_j}\times axis^b_j \cdot c_t + k_m \cdot axis^b_j) } \cdot A_j  \\ & = \sum^{K-1}_{j=0} m_j A_j \label{eq:MOMENTS} 
		\end{split} \\
		A_j &= {\color{black}\bar{ \omega}^j_r \cdot \bar{\omega}^j_r},
	\end{align}
\end{subequations}
where $\bar{\omega}^j_r$ is $j$th $( 0 \leq j \leq K - 1)$ rotor's RPM or angular speed, $c_t$ is the thrust coefficient, and $k_m$ is the moments coefficient. And $axis^b_i$ and $P^b_{r_j}$ are $i$th rotor's unit axis vector and position at the body frame. In addition, the trust model depends on $i$th UAV actuator. Furthermore, we adopt the quadratic model $A_j$ of brushless motors. 

Then, the control allocation formula can be summarised as follows:
\begin{equation}
	\begin{aligned}
	g(u) = \begin{bmatrix}
		T_b \\
		M_b 
	\end{bmatrix}  & =  \begin{bmatrix}
		f_0 & ... & f_{K-1} \\
		m_0 & ... & m_{K-1}
	\end{bmatrix} \begin{bmatrix}
		A_0(\bar{\omega}^0_r) \\ \vdots \\ A_{K-1}(\bar{\omega}^{K-1}_r)
	\end{bmatrix}, \\
	u & =[\bar{\omega}_r^0, ... ,\bar{\omega}_r^{K-1}]^T. 
	\end{aligned} \label{eq:CAF}
\end{equation}

A typical quadrotor's control allocation model is as follows:
\begin{equation}
	g^*(A) = 
	\begin{bmatrix}
	0  & 0 & 0 & 0 \\
	0 & 0 & 0 & 0\\
	c_t & c_t & c_t & c_t\\
	0 & 0 &  l c_t & -l c_t\\
	-l c_t & l c_t & 0 & 0 \\
	k_m & k_m & - k_m & - k_m
	\end{bmatrix} \begin{bmatrix}
		A_0 \\ A_1 \\ A_2 \\ A_3
	\end{bmatrix} = T_e A ,
\end{equation}
where $l$ is the length of every arm of a quadrotor. Theoretically, the quadrotor cannot generate thrust in the x-axis and y-axis at the body frame. 

\subsection{MPC Representation based on FGO}
The general MPC formulation is as follows:
\begin{equation}
	\begin{aligned}
		 \min\limits_{u_{0:N-1}} = & \sum^{N-1}_{k=0} (x_k W_k z_s^k + u_k^T Y_k u_s^k) + x_N W_N z_s^N,  \\
	 	s.t.\ &x_{k+1} = f(x_k) , \ 0 \leq k \leq N-1, \ x_0 = x_{init}, \\
	 	 & u_{min} \leq u_k \leq u_{max}, 
	\end{aligned} \label{eq:GeneralMPC}
\end{equation}  
where $W_k =H_k^TH_k \geq 0 $ and $W_N \geq 0$ are the real symmetric matrices and $Y_k > 0$ is a symmetric real matrix. $u_s$ and $u_s^k$ are input vector and input set-point vector, respectively. $z_s^k$ and $z_s^N$ are state set-point. 

In the field of UAV trajectory tracking, MPC's objective function is equivalent to the sum of squares of multiple error terms, including distance and input constraints for a period of time in the future. It naturally conforms to the factor graph's structure. The distance error is the difference between the predicted trajectory point and the corresponding reference point. The input constraints include the bound limit and rate limit. Although the input $u$ can be generated by the trajectory planning method. We just consider them as the initial value when conducting the FGO's iterative solver. This is because the error between solved input and reference input may not be approximated as Gaussian distribution. But, the input's rate $(u_i  - u_{i+1})$ of actuators is restricted due to its limited capability. Furthermore, the input should confront to actuator's bound limit. 

As a result, (7) can be transformed into a formula for factor graph representation:
\begin{equation}
	\begin{aligned}
		 \min\limits_{u_{0:N-1}} = \{ & \sum^{N-1}_{k=1} \Vert x_k - z_r^k \Vert ^2_{Q_k}  +   \Vert x_N - z_r^N \Vert ^2_{Q_N} \\
		  &  + \sum^{N-2}_{t=0} \Vert u_t- u_{t+1} \Vert ^2_{R_t} \},  \\
	 	s.t.\ &x_{k+1} = f(x_k) , 0 \leq k \leq N-1, x_0 = x_{init}, \\
	 	 & u_{min} \leq u_t \leq u_{max},
	\end{aligned}\label{eq:MPC}
\end{equation}
{\color{black}where $Q_k$, $Q_N$, and $R_t$ represent the weighting matrix.} 

\subsection{Unified FGO Representation}

We try to combine positioning and control into a one-stage problem. The unified optimization problem combining positioning factor, LiDAR factor, and control-related factor is as follows:
\begin{equation}
	\min\limits_{\begin{array}{ll}u_{i:i+N-1}, \\ x_{i-W+1:i}\end{array} }  \left\{ \begin{array}{ll} \sum^{i}_{j = i - W + 1}( { \Vert x_j - z_m^j \Vert _{P_j}^2} \\ + \Vert r^S_j(S_j,\delta x_j) \Vert^2_{p_j}  )\\
	 + \sum^{N-1}_{k=1} \Vert x_{k+i} - z_r^{k+i} \Vert ^2_{Q_k}  \\ +   \Vert x_{N+i} - z_r^{N+i} \Vert ^2_{Q_N}  \\ 
	 + \sum^{N-1}_{l=0} (\Vert f(x_{l+i}) - x^{l+i+1} \Vert ^2_{D_l} \\+ \Vert h(u_{l+i}) \Vert^2_{Q_{lim}} ) \\
	 + \sum^{N-2}_{t=0} \Vert u_{t+i} - u_{t+i+1} \Vert ^2_{R_t} \end{array} \right \}, 
 \label{eq:JECP}
\end{equation}
where $i$ is the current timestamp, and $z^j_m$ represents absolute measurements (e.g. GNSS measurements), and its covariance matrix is $P_j$ at timestamp $j$ ($ i-W+1 \leq j \leq i $). And $S_j$ represents external sensors' measurements (e.g. LiDAR measurements), and its covariance matrix is $p_j$. Noting that the sensor's error model $r_j^S$  depends on itself. $W$ is the state length of positioning part. 
 
Additionally, the control part involves the reference trajectory factor, dynamics control factor, control limit factor (CLF), and input constraint factor. The input's rate constriction can be added into the factor graph through between factors. In addition, the dynamics control factor establishes constraint between two successive states from the current state $x_i$ to the terminal state $x_{N+i}$. The remaining terms about the predicted states are the reference trajectory factor, in which $z_r$ represents the trajectory's reference point. Finally, $h(u)$ and $Q_{lim}$ represent the control limit factor's error function and its covariance.

\section{JOINT POSITIONING AND CONTROL}
 
\begin{figure}[ht]
      \centering
      \includegraphics[width=0.9\linewidth]{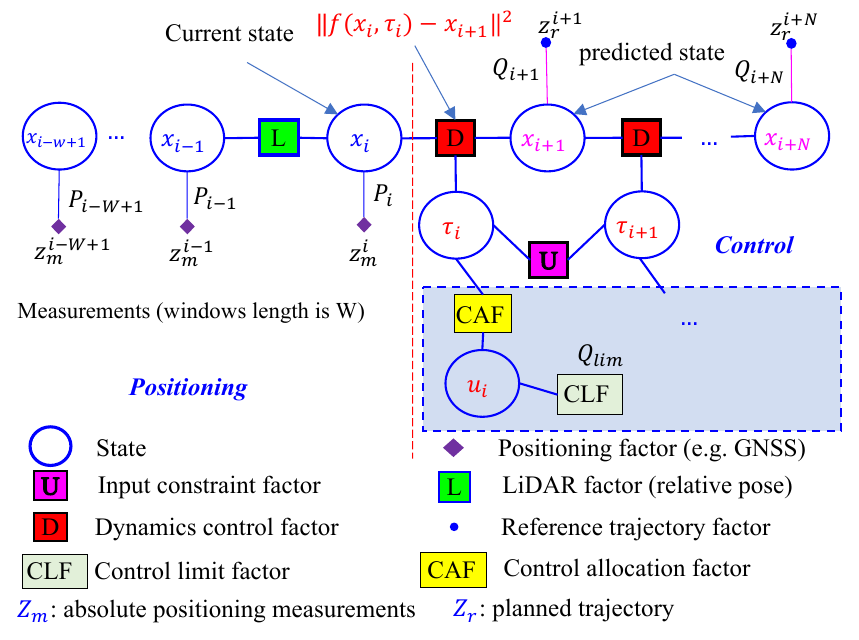}
      \caption{Unified factor graph based on sliding window.}
      \label{img:factor_graph}
 \end{figure}
 
 We provide a unified factor graph framework to achieve JPCM. As illustrated in Fig. \ref{img:factor_graph}, the unified factor graph slides over a time window, which contains $W$ states and $N$ predicted states. The yellow part is the positioning part which includes positioning measurement factors and LiDAR factors (relative pose). Besides, the control part contains the dynamics planning factor, reference trajectory factor, input constraint factor, control allocation factor (CAF), and control limit factor. The dynamic planning factor establishes dynamics constraints on the states of adjacent moments. The reference trajectory factor propels the predicted trajectory to the planned reference trajectory. Moreover, the input constraint factor and control limit factor ensure that the input's solution complies with the driving ability of the motor. 
 
It should be noted that in the application of the JPCM on UAVs, we design a new factor (CAF) based on its dynamic equation to connect the thrust, moments, and the control input of the motor. Besides, the input constraint factor applies to the thrust and moments, limiting their rate of change. 

\subsection{Measurement Factors} 
In actual positioning systems, GNSS provides positioning information and integrates it with inertial measurement unit (IMU). For simplicity, we assume that the measurements of the positioning factor include attitude, velocity, and angular velocity in the simulation. Hence, the positioning factor's model is as follows:
\begin{equation}
	\widehat z_m = \overline{z}_m + n_a, n_a \sim N(0,P),\  z_m = [ p^w_b, R^w_b, v^w_b, \omega_b]^T.
\end{equation}

Therefore, we use multivariate Gaussian measurement factors for state measurements:
\begin{equation}
	\phi _{x_j}^{} \varpropto exp \{ \frac{1}{2} \Vert x_j - z^j_m \Vert ^2_{P_j} \}.
\end{equation}

In addition, we simulate the relative pose transformation $T_{L_i}^{L_j}$  (the $j$th  body's pose at $i$th body frame), which can be computed by iterative closest point (ICP) method toward LiDAR's point cloud. Hence, the LiDAR factor is simplified as: 
\begin{equation}
	\phi _{LiDAR_j}^{} \varpropto exp \{ \frac{1}{2} \Vert T^{L_i}_{L_j} \cdot (T^w_{L_j})^T \cdot T^w_{L_i} \Vert ^2_{^L P^i_j} \},
\end{equation}
where $T_{L_i}^w$ represent the LiDAR's pose at timestamp $i$, $T_{L_j}^w$ represent the LiDAR's pose at timestamp $j$, and $^L P^i_j$ is the covariance matrix of relative pose  $T^{L_i}_{L_j}$. 

\subsection{Dynamics Control and Reference Trajectory Factor}
The discrete form of state propagation is $x_{i+1} = x_i + \dot{x_i} \delta t $.  Thus, the error term can be represented as $e_D=[e_p, e_v, e_\theta, e_\omega]^T = x_{i+1} - x_i - \dot{x_i} \delta t $. According to (\ref{eq:CLD_Dyn}), the dynamics control factor's errors are as follows:
\begin{subequations}
	\begin{align}
		e_p & = p^w_{b_{i+1}} - v^w_{b_i} \cdot \delta t - p^w_{b_i} \\
		e_v & = v^w_{b_{i+1}} - v^w_{b_i} - (-e_3 g + R^w_{b_i} T_{b_i}  / m_b) \cdot \delta t \\
		e_\theta & = Log((R^w_{b_{i+1}})^{-1}R^w_{b_i}(I + \lfloor \omega_{b_i} \times \rfloor \cdot \delta t) )\\
		e_\omega & = \omega_{b_{i+1}} - \omega_{b_i} - I^{-1}_b( M_{b_i}-\omega_{b_i} \times I_b \omega_{b_i}) \cdot \delta t,
\end{align}\label{eq:DCF}
\end{subequations}
where $\delta t$ is period time between two continuous states. And $Log(*)$ and $Exp(*)$ represent and Logarithmic map and Exponential map of Lie theory, respectively.

In addition, most Jacobian matrices are easy to be derived. And only essential Jacobian matrixes are given. The Jacobian matrix of rotation $e_\theta$ error with angular speed $\omega$ is as follows: 
\begin{equation}
\begin{aligned}
		\frac{\partial e_\theta }{\partial \omega_{b_i}} & = \frac{\partial {Log}( (R^w_{b_i+1})^{-1} R^w_{b_i} {Exp}(\omega_{b_i}  \cdot \delta t ) )}{\partial \omega_{b_i}}  \\
		& =  \frac{\partial {Exp}(\omega_{b_i} \cdot \delta t)} {\partial \omega_{b_i}}  \\
		& = {dexp} \cdot \delta t,
\end{aligned} \label{eq:JAC_THETA_OMEGA}
\end{equation}
where $dexp$ is given in the appendix section.

Moreover, the Jacobian matrix $e_\omega$ with $\omega_{b_i}$ is as follows:
\begin{equation}
\begin{aligned}
		\frac{\partial e_\omega }{\partial \omega_{b_i}} & = \frac{\partial (-\omega_{b_i} + I^{-1}_b \omega_{b_i} \times I_b \omega_{b_i} \delta t)}{\partial \omega_{b_i}} \\
		& = -I_3 +\begin{bmatrix}
		0 & a \omega_3 & a \omega_2 \\ 
		b \omega_3 & 0 & b \omega_1 \\ 
		c \omega_2 & c \omega_1 & 0 
			\end{bmatrix} \delta t,
	\end{aligned} 
	 \label{eq:JAC_OMEGA_OMEGA}
\end{equation}
where $\omega_1, \omega_2, \omega_3$ represents vector $\omega_b$'s components. And $a$, $b$, $c$ are as follows:
\begin{equation}
	\begin{aligned}
		a &= (I^1_b)^{-1}(I^3_b - I^2_b) 	\\
		b &= (I^2_b)^{-1}(I^1_b - I^3_b) \\
		c &= (I^3_b)^{-1}(I^2_b - I^1_b).
	\end{aligned}
\end{equation}
 \begin{figure*}[htbp]
      \centering
      \includegraphics[width=1.0\linewidth]{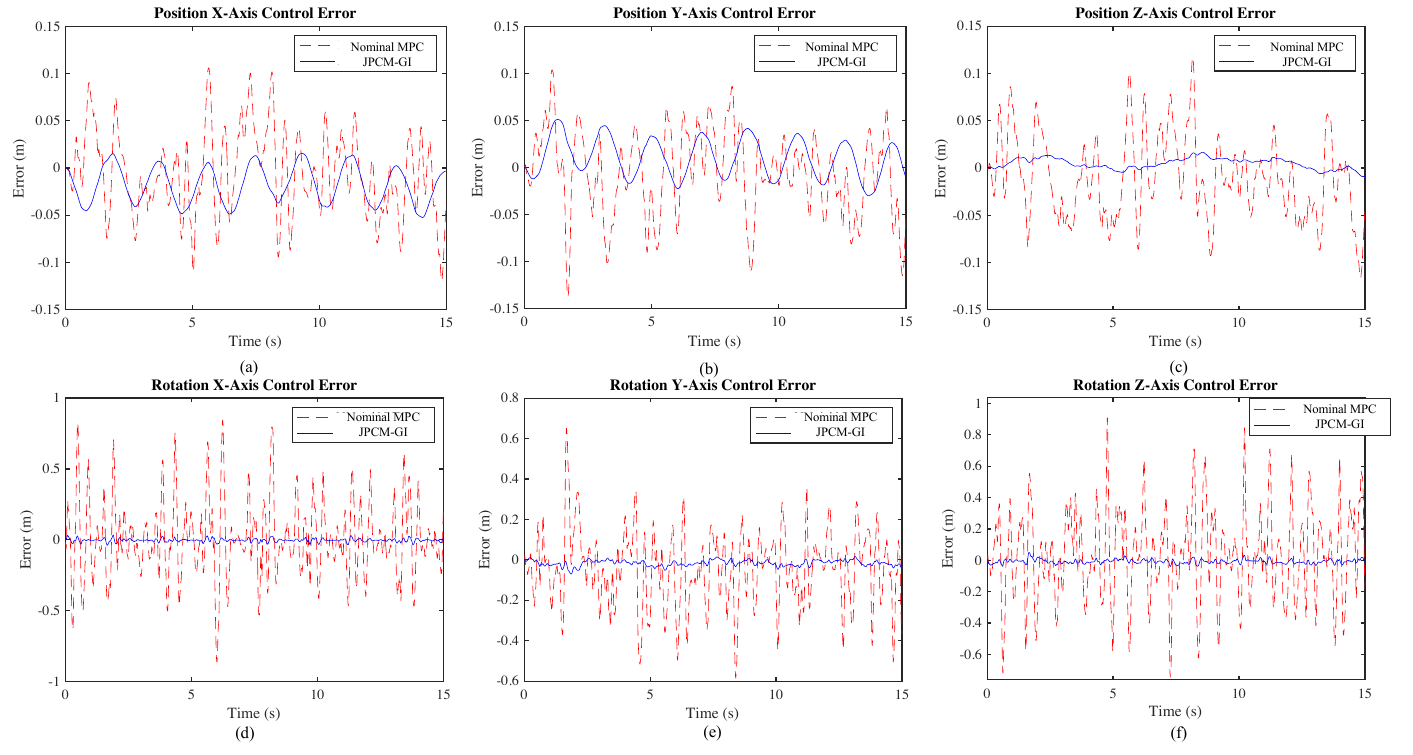}
      \caption{Position and rotation control errors.}
	 \label{img:PositionRotationError}
\end{figure*}
The reference trajectory factor is a between factor. Its error is as follows:
\begin{equation}
e_r = (p,R,v)  \ominus (p_r, R_r, v_r),
\end{equation}
where the position and velocity reference error can be obtained by direct subtraction, and the rotation error is ($R^T R_r$).

In addition, the reference trajectory factor's covariance $Q_k ( 1 \leq k \leq N)$ can be configured by users. 

\subsection{Control Allocation Factor (CAF) and Control Limit Factor (CLF)}
The quadrotor is a kind of widely used UAV that installs four propellers with the same axial direction and symmetrical distribution. However, the CAF handles freely designed geometry of multi-rotor UAVs. The error formula of CAF is as follows:
\begin{equation}
	e_{ACF} = \tau - g(u). 
\end{equation}
                    
The CLF is designed to ensure that control trajectories confront to actuator's limits.  Defining the hinge loss cost function for the $u$ control input inequality, which is as follows \cite{SCTEforCollision}: 
\begin{equation}
	h(u_i) = \left\{  \begin{array}{ll}
		u^j_{min} + u^j_{ths} - u^j_i  & if\ u^j_i < u^j_{min} + u^j_{ths}, \\
		u^j_i- u^j_{max} + u^j_{ths}  & if\ u^j_{max} - u^j_{ths} \leq u_i^j , \\
		0 & otherwise,
	\end{array}
	\right. 
\end{equation}
where $u^j_i$ is $j$th component of input $u_i$, $u^j_{min}$ is input lower bound, and $u^j_{max}$ is input upper bound, and $u^j_{ths}$ is a threshold value. If the value is within the threshold, then the cost is 0. Hence, limit violations are ensured during the optimization. In addition, $Q_{lim}$ is a CLF's covariance matrix which determines how fast the error grows as the value approaches the limit. 

\section{SIMULATION RESULTS}

The quadrotor's trajectory control simulation is conducted to evaluate the proposed JPCM's performance. And the reference trajectory is assumed a constant linear velocity circle. The trajectory is generated by the minimum snap trajectory generation method \cite{MiniSnap}. Moreover, the Gaussian white noise is added to the thrust and angular speed of the quadrotor simulator.  The gravity of the quadrotor is approximately $10N$. The thrust noise's sigma is $1N$, and the angular speed noise's sigma is $0.02rad/s$. The circle trajectory's linear speed is 5m/s, and its radius is 1.5m. Finally, the extrinsic parameter between the UAV body frame and the LiDAR frame is set to an identity matrix. The proposed unified factor graph is built based on the open-source software GTSAM \cite{GTSAM}. 

\subsection{The Comparison between the Nominal MPC and JPCM}
In the simulation, we don't simulate the IMU pre-integration process. Therefore, we assume that the positioning factor's measurements also include attitude, velocity, and angular speed. 

To evaluate the performance, it is assumed that only current positioning measurements exist ($W=1$). And the GNSS position's sigmas, rotation's sigmas, velocity's sigmas and angular velocity's sigma are configured as $0.20m$,  $0.05m/s$, $0.01rad/s$, and $0.001rad/s$, respectively.  It is worth noting that the nominal MPC is solved by FGO, in which the positioning results are fed into MPC as the initial state's value $x_0$. 

The JPCM's parameters setting is as follows. The circle trajectory's linear speed is 5m/s, and its radius is 1.5m.  The predicted states length $N=20$, the reference state covariance matrix $Q_k$ = $diag(0.03^2, 0.03^2, 0.03^2, 0.3^2, 0.3^2, 0.3^2, 3^2, 3^2, 3^2)$ and $Q_N$ = $diag(0.005^2$, $0.005^2$, $0.005^2$, $0.3^2$, $0.3^2$, $0.3^2$, $3^2$, $3^2$, $3^2)$, the covariance matrix $R_l$ of input constraint factor is $diag(1.0, 0.5, 0.5, 0.5)$. The nominal MPC's parameters are consistent with JPCM. 

 \begin{figure}[htbp]
      \centering
      \includegraphics[width=0.85\linewidth]{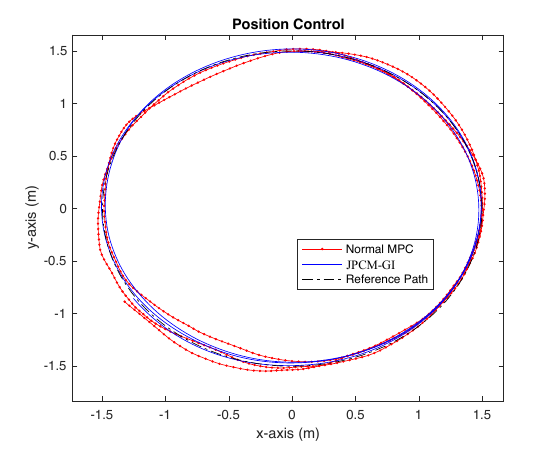}
      \caption{The paths of Nominal MPC and JPCM-GI. The black dashed line, red solid line, and blue solid line represent the reference path, the tracking path based on nominal MPC, and the tracking path based on JPCM-GI, respectively. }
	 \label{img:CirclePosition}
\end{figure}

As shown in Fig. \ref{img:CirclePosition},  the path of JPCM-GI (JPCM with only current single positioning measurement) fits the reference path better than the nominal MPC method. Conversely, nominal MPC shows severe jitter. In addition, its rotation and position control errors are illustrated in Fig. \ref{img:PositionRotationError}, In addition, the Root-mean-square errors (RMSEs) are summarised in Table. \ref{tab:RMSE}. It is readily apparent that the MPC problem solved by FGO (MPC-NL) converges to the reference trajectory with very small control errors when there is no noise in the positioning measurements. However, compared with MPC-NL, nominal MPC's control errors significantly increase. It demonstrates that noisy positioning has a deteriorating impact on UAV's control performance. Furthermore, the table shows that the position control errors and rotation control errors of JPCM-GI decrease compared to nominal MPC with noise positioning. 

In summary, when accurate positioning is available, the MPC control performance based on FGO is excellent. However, in urban applications, severe positioning noise is often present, which deteriorates the MPC control error. On the other hand, the proposed JPCM considerably improves trajectory tracking performance in urban scenes. The JPCM achieves a more robust performance. 

\begin{table}[htbp]
    \caption{The RMSE of MPC-NL (nominal MPC with noiseless positioning measurements), Nominal MPC, and JPCM-GI. \label{tab:RMSE}}
    \centering
   \resizebox{1.0 \linewidth}{!}{
    \begin{tabular}{ccccccc}
    \toprule
    Error & \multicolumn{3}{c}{Position} & \multicolumn{3}{c}{Rotation} \\
      & (m) & (m) & (m) 	& (rad) 	& (rad) 	& (rad)  \\
    \midrule
    MPC-NL        &  0.0075  &  0.0072 &  0.0036   &  0.0049  &  0.0051  &  0.0036 \\
 	Nominal MPC       &  0.0434  &  0.0451  &  0.0453  &  0.2717  &  0.2007 & 0.2513 \\
    JPCM-GI &  0.0263  &  0.0229  &  0.0073  &  0.0116  &   0.0218 & 0.0153 \\
    \bottomrule
	\end{tabular}}
\end{table}

\subsection{Unplanned Movements Recovery}
\begin{figure}[htbp]
    \centering
  \subfloat[\label{img:rec1a}]{%
       \includegraphics[width=0.9\linewidth]{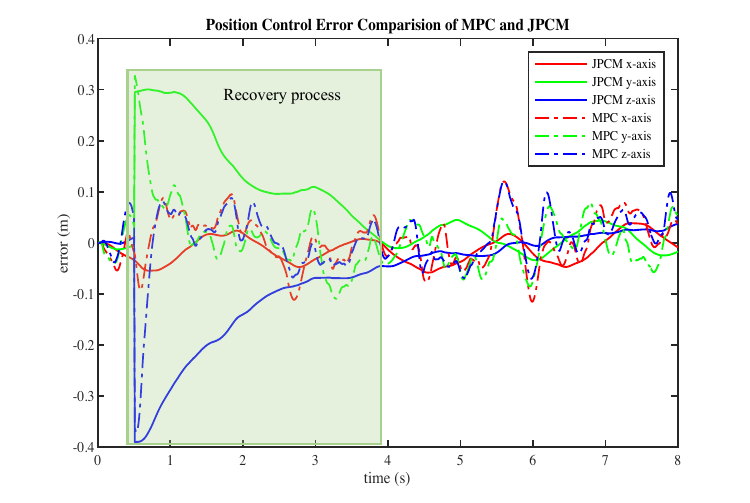}}
    \\
  \subfloat[\label{img:rec1b}]{%
        \includegraphics[width=0.9\linewidth]{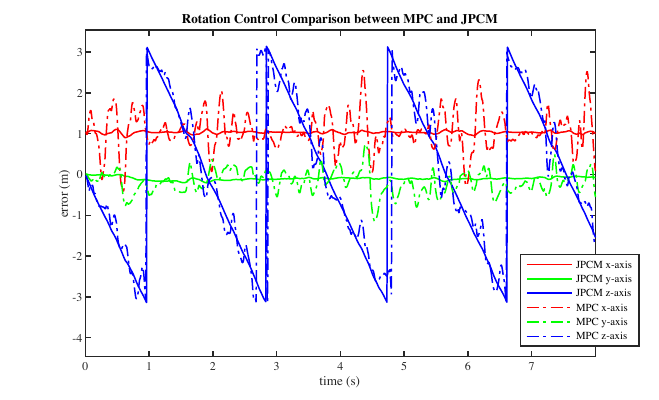}}
      \caption{The JPCM-GI-based position control recovery process under an unplanned movement ($\Delta p_b^w  = [0.00m,0.30m,-0.40m]^T$). (a) The position control error comparison of MPC and JPCM. (b) The attitude control comparison of MPC and JPCM. }
	 \label{img:PositionErrRec}
\end{figure}

There is a situation where, for example, a brief shutdown of the engine or manual pushing, can cause significant unplanned movement. The controller should have the ability to maintain a stable status. The reference state covariance matrix $Q_k$ is configured as $diag(0.2^2, 0.2^2, 0.2^2, 0.3^2, 0.3^2, 0.3^2, 3^2, 3^2, 3^2)$ , and $Q_N$ is not changed. As shown in Fig. \ref{img:rec1a}, an unplanned movement $\Delta p_b^w  = [0.00m,0.30m,-0.40m]^T$ is simulated in 0.5 seconds, and it can be seen that the UAV returns to its planned trajectory in about three seconds. Compared with MPC, the JPCM's recovery process takes longer, but the position error curve is more stable. Besides, the attitude's jitter is significantly suppressed as shown in Fig. \ref{img:rec1b}. 

\subsection{JPCM Based on Sliding Windows (SW-JPCM) Verification}
The JPCM based on sliding window simulation is conducted with $W = 10$. The simulation parameters and JPCM parameters are the same as the previous configuration. The extrinsic parameter is an identity matrix. In addition, the LiDAR's measurement frequency is 100Hz. The covariance of the pose increments of adjacent frames, i.e. $^L P^i_j$, is designed as $diag(0.001^2, 0.001^2, 0.001^2, 0.001^2, 0.001^2, 0.001^2)$. 

\begin{figure}[htbp]
    \centering
  \subfloat[\label{1a}]{%
       \includegraphics[width=0.9\linewidth]{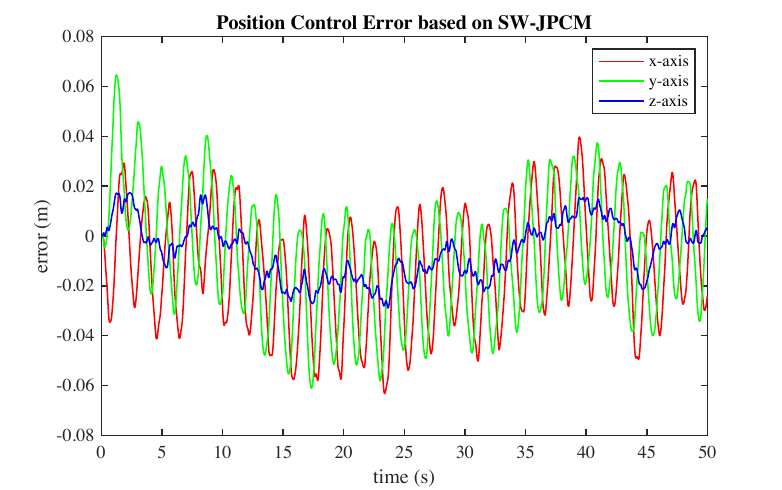}}
    \\
  \subfloat[\label{1b}]{%
        \includegraphics[width=0.9\linewidth]{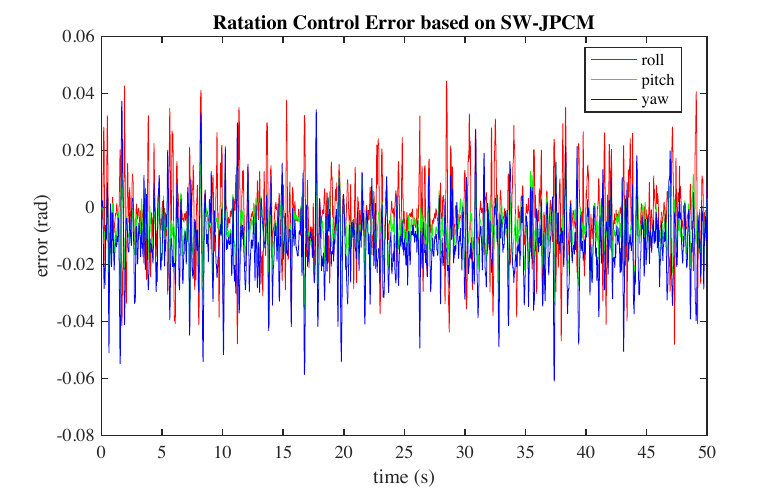}}
\caption{The control errors of SW-JPCM (W = 10). (a) The position control errors. (b) The attitude control errors. }
      \label{img:SW-JPCM}
\end{figure}

As shown in Fig. \ref{img:SW-JPCM}, the position and rotation control error curves demonstrate that SW-JPCM is feasible to actualize joint positioning and control. In addition, the control error is effectively converged. 

\section{DISCUSSION}

\begin{figure}[htbp]
      \centering
      \includegraphics[width=0.6\linewidth]{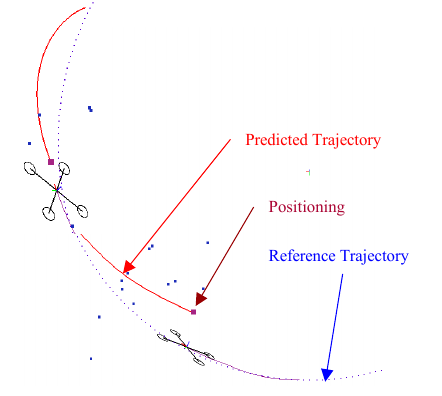}
      \caption{The predicted trajectory jitter due to state mismatch. }
      \label{img:Fluctuation}
\end{figure}

When the initial state error is large, it will cause significant fluctuations in the predicted trajectory as to traditional MPC as depicted in Fig. \ref{img:Fluctuation}. Because the solution of dynamic control does not match the real position of the actual UAV, this may lead to its unstable control. In addition, the model of UAV is highly nonlinear, and the attitude jitter will pose a significant threat to safety.  Conversely, the control considering initial positioning probability may break its bottleneck. It can be seen that considering probability can decrease the variance of errors and result in a more stable performance. This is because representing positioning through probability can reduce the risk of control overshoot. Compared with the nominal MPC, the JPCM can relax the hard constraint with initial state in general MPC optimization problem. 

Additionally, the recovery process shows that the JPCM needs more time to get back to the planned path compared with nominal MPC. This is because when optimizing positioning and control simultaneously, due to the constraints of the planned trajectory, the estimated current state is pulled closer to the trajectory, reducing the energy of control. Traditional MPC can plan actions without hesitation, which also makes it more prone to jitter. 

\section{CONCLUSIONS}

To break the information gap between positioning and control, a joint positioning and control method is proposed. The JPCM combines positioning and control into a unified factor graph. Additionally, a design framework for the unified factor graph is also provided. Especially in the trajectory control part, we provide factors related to dynamic control and trajectory tracking. Finally, a quadrotor simulator is used to evaluate the proposed method's performance. The simulation results show that the proposed method is convergent with smaller errors than nominal MPC. For future research, we will introduce IMU simulation into the system. 

\addtolength{\textheight}{-12cm}   



\section*{APPENDIX}  \label{sec:APPENDIX}
For every 3-vector $\omega$ there is a corresponding rotation matrix $Exp(\omega) = exp(\lfloor \omega \times \rfloor )$. It is equivalent to the axis-angle representation for rotations, where the unit vector $ \omega / \theta$ defines the rotation axis, and its magnitude the amount of rotation $\theta$. 
\begin{equation}
	\begin{aligned}
		dexp &=  \frac{\partial {Exp}(\omega)} {\partial \omega} \\
		& =  \frac{\partial cos(\theta)I_3 + \frac{sin (\theta)}{\theta}  \lfloor \omega \times \rfloor +   \frac{1-cos (\theta)}{\theta^2} \omega \omega^T } {\partial \omega}  \\
		& = I_3 - a \times K + b \times K^2,
	\end{aligned}
\end{equation}
where $a = 2  \frac{sin^2(\theta/2)}{\theta}$, $b = 1 -  \frac{sin(\theta)}{\theta}$, and $K =   \frac{\lfloor \omega \times \rfloor }{\theta}$.  If $\omega$ nears zero, then the $dexp$ approximates $I_3 - 0.5 \lfloor \omega \times \rfloor$.


\bibliographystyle{./IEEEtran} 
\bibliography{./References}

\end{document}